\documentclass[runningheads]{llncs}
\let\subparagraph\paragraph

\usepackage[T1]{fontenc}
\input{paper.sty}

\title{Towards a Framework for Comparing the Complexity of Robotic Tasks}
\titlerunning{Comparing the Complexity of Robotic Tasks}  

\author{Michelle Ho\thanks{Equal contribution.} \and Alec Farid$^\star$ \and Anirudha Majumdar}
\authorrunning{M. Ho, A. Farid, and A. Majumdar} 

\institute{Dept. of Mechanical and Aerospace Engineering, Princeton University \\
\email{\{mtho, afarid, ani.majumdar\}@princeton.edu}}

\begin{document}
\mainmatter              
\maketitle
\vspace{-1.2em}
\begin{abstract}
We are motivated by the problem of comparing the \emph{complexity} of one robotic task relative to another. To this end, we define a notion of \emph{reduction} that formalizes the following intuition: Task~1 reduces to Task~2 if we can efficiently transform \emph{any} policy that solves Task~2 into a policy that solves Task~1. We further define a quantitative measure of the \emph{relative complexity} between any two tasks for a given robot. We prove useful properties of our notion of reduction (e.g., reflexivity, transitivity, and antisymmetry) and relative complexity measure (e.g., nonnegativity and monotonicity). In addition, we propose practical algorithms for estimating the relative complexity measure. We illustrate our framework for comparing robotic tasks using (i) examples where one can analytically establish reductions, and (ii) reinforcement learning examples where the proposed algorithm can estimate the relative complexity between tasks. 
\end{abstract}
% \vspace{-10pt}

\section{Introduction}
\vspace{-5pt}

Consider the following pairs of robotic tasks: (i) an autonomous truck driving on one side of the road vs. the other side, (ii)  an autonomous car driving in one city vs. another, and (iii) a cartpole balancing the pole upright  vs. downward.
Intuitively, the two tasks in (i) are ``as hard as each other'', while one task in (iii) (i.e., balancing upright) is more challenging than the other (i.e., balancing downward).
The tasks in (ii) may be challenging in different ways and thus may not admit a straightforward ordering. How can we \emph{formally} compare the \emph{complexity} of each pair of tasks?

Despite decades of algorithmic advancements in robotics, we currently lack precise mathematical foundations for answering such questions. In contrast, \emph{computational complexity theory} \cite{Arora09} provides a unifying framework and set of tools for establishing and comparing the difficulty of \emph{all} computational problems. It also guides practitioners by setting expectations for the kinds of algorithms that will solve a given problem. For example, a problem that is as hard as 3-SAT will not admit a polynomial-time solution (unless P = NP); thus, a practitioner is motivated to find \emph{approximation algorithms} for such a problem. We currently lack such a unifying theory for understanding the complexity of robotic tasks.
% There are currently no clear candidates for such a unifying theory in robotics. 

{\bf Statement of Contributions.} Motivated by this challenge, we take a step towards developing a precise framework for establishing the \emph{relative complexity} of robotic tasks. To this end, we make five specific contributions. 
\begin{itemize}
    \item {\bf Reductions between tasks.} Our key insight is to define a notion of \emph{reduction} \cite[Ch. 2]{Arora09} between robotic tasks (Definition~\ref{def:task reduction}). This definition formalizes the following intuition: Task~1 reduces to Task~2 if we can transform \emph{any} policy that solves Task~2 into a policy that solves Task~1. The reduction is ``easy'' if the transformation is computationally efficient (e.g., polynomial-time). Crucially, this notion of reduction captures the relative complexity of tasks in terms of the complexity of \emph{online} task execution (i.e., how challenging the two tasks are from the perspective of the robot rather than the robot's software designer). 
    \item {\bf Quantifying relative complexity.} We propose a \emph{quantitative measure} (that outputs values in the range $[0,1]$) for comparing the complexity of one robotic task relative to another (Definition~\ref{def:relative complexity}). One can think of this measure as a ``smoothed'' version of our notion of reduction, where this notion of relative complexity can be defined for arbitrary tasks (in contrast to the definition of reduction, which captures a strict and binary notion of relative complexity).
    \item {\bf Properties of reduction and relative complexity.} We prove basic properties of our notion of reduction between tasks (e.g., reflexivity, transitivity, and antisymmetry in Proposition~\ref{prop:task reduction}) and the relative complexity measure (e.g., non-negativity and monotonicity in Proposition~\ref{prop:relative complexity}). 
    \item {\bf Algorithm.} We propose a practical algorithm based on \emph{adversarial training} for estimating the relative complexity measure for robotic tasks in reinforcement learning contexts (Algorithm~\ref{alg:relativecomplexity}). 
    \item {\bf Examples.} We demonstrate our framework using (i) illustrative examples where one can analytically establish reductions (Sec.~\ref{sec:examples of reductions}), and (ii) numerical examples based on reinforcement learning problems where we apply our proposed algorithm for estimating relative complexity (Sec.~\ref{sec:examples}).  
\end{itemize}

\section{Related Work}
\vspace{-5pt}
\label{sec:related work}

{\bf Complexity and robotics.} 
The study of complexity theoretic questions in robotics has a long history. Early work \cite{Reif79, Canny88} established the PSPACE-complete-ness of the general motion planning problem. Other results include PSPACE-hardness \cite{Hopcroft84, Hopcroft84a, Joseph85, Culberson97, Solovey16} and NP-hardness \cite{Murrieta08, Borie09,Borie11, Hauser14, Han17}  of various planning problems. Complexity results for systems with nontrivial (e.g., nonlinear or uncertain) dynamics have also been explored in control theory \cite{Blondel00, Ahmadi13}. The problem we consider in this paper is fundamentally different from the ones above. Specifically, the model of computation in the problems mentioned above assumes that the Turing machine is provided with a complete encoding of the problem on its tape at the very outset (e.g., rational numbers encoding the linear inequalities that define polytopic obstacles in the environment); the algorithm's task is then to perform a given computation on this fixed encoding. In contrast, we are interested in the computational resources required by a robot \emph{as it is performing a given task}. In this model, the robot's embodiment is of crucial importance; instead of a fixed encoding of the input, the robot's sensors provide information incrementally and interactively based on its control actions and the environment.

{\bf Reductions for comparing robots and sensors.}  
% A formalism for measuring the intrinsic complexity of robotic tasks in terms of \emph{information invariants} was proposed in \cite{Donald95}. 
The pioneering work in \cite{Donald95} proposed a formalism for measuring the intrinsic complexity of robotic tasks in terms of \emph{information invariants} and a notion of reduction for comparing different robots.
% This framework allows one to formally compare the power of different robots via a notion of reduction.
The work presented in \cite{OKane08} has a similar goal, but employs the notion of \emph{information spaces} \citep[Chapter 11]{Lavalle06} in order to handle tasks where sensors only provide partial state information. Notions of reductions for comparing the power of different sensors have also been developed \cite{Lavalle12, Erdmann95}. While our work is inspired by the frameworks above and also uses the idea of reductions, our focus is distinct. In particular, our goal is to compare different robotic \emph{tasks} (instead of different robots). In addition, we propose a measure of \emph{relative complexity} between two tasks that goes beyond the strict notion of reduction and allows one to quantify how complex one task is with respect to another in terms of online computational resources required by the robot. 

{\bf Reductions and task complexity in learning.} 
Our definition of reduction between tasks formally captures the intuition that Task 2 is at least as complex as Task 1 if any solution (in the form of a control policy) for Task 2 can be transformed into a solution for Task 1. The idea of transforming solutions for one task into solutions for another has also been exploited for tackling problems in machine learning. In particular, \cite{Chang19, Li21} propose approaches for composing previously-learned ``modules'' in order to speed up the process of learning on a new problem. In contrast to our work, these approaches do not seek to compare the difficulty of tasks (their goal is to obtain practical benefits in terms of sample efficiency and speed of learning). There has also been work on defining (asymmetric) notions of distance between supervised learning tasks \cite{achille19, achille20, Tran19} (e.g., using techniques from Kolmogorov complexity theory \cite{Li08}). These approaches are motivated by problems in \emph{transfer learning} and attempt to capture how quickly solutions from one learning problem can be fine-tuned for a new learning problem. In contrast to these measures, we seek to capture the relative complexity between tasks in terms of the complexity of computations that must be executed \emph{online} (instead of the complexity of offline fine-tuning). In addition, we are motivated by robotic tasks in contrast to supervised learning problems.  
\section{Problem Formulation}
\vspace{-5pt}

\label{sec:problem formulation}
\textbf{Robots, Environments, and Rewards.} Let $p(s_{t} | s_{t-1}, a_{t-1})$ describe a robot's dynamics, where $s_t \in \mathcal{S}$ represents the combined state of the robot and its environment at time-step $t$, and $a_t \in \A$ represents the action. Let $p_0$ denote the initial state distribution. We denote the robot's sensor mapping as $\sigma(o_t|s_t)$. We consider robotic tasks that are prescribed using reward functions; let $\sum_{t} r(s_t, a_t) \in \RR$ denote the cumulative reward over a given (finite or infinite) time horizon. 

\textbf{Policies.} Let $\pi: \O \rightarrow \A$ denote a policy for some observation space $\O$ and action space $\A$. We will denote the set of all policies (i.e., all mappings from $\O$ to $\A$) by $\Pi$. One can extend the formulation we present to policies with memory by augmenting the observation space to keep track of memory states (potentially in a similar way to \cite{Saberifar19, OKane20}). We will define the reward achieved by a policy as:
\begin{equation}
    R(\pi) \coloneqq \min\bigg(\underset{p_{0}, p, \sigma}{\EE} \sum_{t} r(s_t, \pi(o_t)), R^\star\bigg),
\end{equation} 
where $R^\star$ is a chosen success threshold and forms an upper bound on $R(\pi)$.

{\bf Tasks.} A \emph{task} is formally defined as a partially observable Markov decision process (POMDP) using the tuple $\t := (\S, \A, \O, p, \sigma, r, p_0, R^\star)$ which describes the state space, action space, observation space, dynamics, sensor, reward function, the distribution over initial states, and the threshold for success. For example, a task $\t$ could correspond to a cartpole swinging up and balancing (with random initial conditions) or a drone navigating through random obstacle environments (where the randomness over obstacle placements is defined using $p_0$; recall that the state $s_t$ encapsulates the combined state of the robot and its environment). We will index tasks using $\xi$ and let $\T = \{ \tau_\xi\}_\xi$ denote a set of tasks. A particular task in $\T$ is thus denoted as $\t_\xi := (\S_\xi, \A_\xi, \O_\xi, p_\xi, \sigma_\xi, r_\xi, p_{0,\xi}, R^\star_\xi)$. We will say that a policy $\pi_\xi$ is \emph{admissible} on task $\t_\xi$ if $R_\xi(\pi_\xi) = R^\star_\xi$. We use shorthand $\pi^\star_\xi$ to denote an admissible policy on task $\t_\xi$. Let $\Pi^\star_\xi \subseteq \Pi_\xi$ be the set of all admissible policies on task $\t_\xi$.

% \textbf{Policies.} Let $\pi_\xi: \O_\xi \rightarrow \A_\xi$ denote a policy for task $\tau_\xi$. We will denote the set of all policies (i.e., all mappings from $\O_\xi$ to $\A_\xi$) by $\Pi_\xi$. One can extend the formulation to policies with memory by augmenting the observation space to keep track of memory states \edit{(potentially in a similar way to \cite{Saberifar19, OKane20})} \alec{double check these citations}. We will define the reward achieved by a policy on a particular task $\t_\xi$ as:
% \begin{equation}
%     R_\xi(\pi_\xi) \coloneqq \min\bigg(\underset{p_{0,\xi}, p_\xi, \sigma}{\EE} \sum_{t} r_\xi(s_t, \pi_\xi(o_t)), R^\star_\xi\bigg),
% \end{equation} 
% where $R^\star_\xi$ is a chosen success threshold for task $\t_\xi$ (which forms an upper bound on $R_\xi(\pi_\xi)$). We will say that a policy $\pi_\xi$ is \emph{admissible} on task $\t_\xi$ if $R_\xi(\pi_\xi) = R^\star_\xi$. We use shorthand $\pi^\star_\xi$ to denote an admissible policy on task $\t_\xi$. Let $\Pi^\star_\xi \subseteq \Pi_\xi$ be the set of all admissible policies on task $\t_\xi$.

\textbf{Task reduction and relative complexity.} Our goal is to introduce notions which meaningfully and quantitatively compare the complexity of robotic tasks. First, we aim to develop a \textit{binary relation} (denoted by ``$\tleq$") between two tasks which will provide a notion of reduction, i.e., if task $\t_1$ reduces to task $\t_2$ ($\t_1 \tleq \t_2$), then task $\t_2$ is at least as complex as task $\t_1$. Second, we aim to formulate a measure $\T \times \T \rightarrow [0,1]$ that compares the relative complexity of one task with respect to another. The goal is to not only establish if one task is more complex than another, but the \emph{degree} to which it is more complex. 
\section{Task Reduction}
\vspace{-5pt}

We propose a definition of \emph{reduction} for robotic tasks in order to formalize what it means for one task to be as hard as another. The idea of reductions comes from the theory of computational complexity; we review basic definitions in Sec.~\ref{sec:background} before describing reductions between robotic tasks in Sec.~\ref{sec:reductions}.  

\subsection{Background}
\label{sec:background}
We begin with the definition of a \textit{Turing} reduction between two computational problems. Let $A$ and $B$ be \textit{decision problems}, i.e., problems where each instance has a yes/no answer (e.g., 3-SAT). Let $\oracle_B$ be an oracle for decision problem $B$. This oracle is a blackbox which outputs the solution to any instance $b \in B$. 

\begin{definition}[Turing Reduction \cite{Sipser96}] Decision problem $A$ reduces to $B$ (written $A \tleq_T B$) if we can compute the solution to all instances of $A$ using an oracle $\oracle_B$ for $B$. 
\label{def:turing reduction}
\end{definition} 
Intuitively, $A$ reduces to $B$ if one can use an oracle for $B$ as a sub-routine for solving instances of $A$, and thus B is at least as complex as A. Note that this definition does not include reference to the complexity of the resulting function for solving $A$. We are specifically interested in \emph{efficient} reductions. A particular notion of efficiency is formalized by \emph{polynomial-time} reductions, as defined below. 
% Note that this definition does not include reference to the complexity of the either decision problem. We are specifically interested in efficiently computable reductions, i.e. \textit{polynomial-time} reductions.
\begin{definition}[Cook Reduction \cite{Sipser96}] Decision problem $A$ is polynomial-time reducible to $B$ (written $A \tleq_T^P B$) if $A \tleq_T B$ 
and the solution to any instance of $A$ makes only a polynomial number of calls to $\oracle_B$. 
\end{definition}

Importantly, if $\oracle_B$ is efficient (i.e., runs in polynomial time) and we have $A \tleq_T^P B$, then the resulting function for solving $A$ is also efficient. We aim to define a notion of reduction for robotic tasks that is inspired by the notions above for computational problems. In particular, we will leverage the core concept of using an oracle for one decision problem in order to solve another.   

\subsection{Task Reduction: Definition and Properties}
\label{sec:reductions}

In this section, we propose a notion of reduction between two robotic tasks and demonstrate that this defines a partial ordering on a given space $\T$ of tasks. Our notion of reduction captures the following intuition: task $\t_1$ reduces to task $\t_2$ if we can utilize a policy (``oracle'') for $\t_2$ in order to solve $\t_1$. In order to formalize this, we first introduce \emph{encoders} and \emph{decoders}. 

\begin{definition}[Encoder and Decoder] Let $\T$ be a space of tasks and let $\t_1,\t_2 \in \T$ be two tasks (as defined in Sec.~\ref{sec:problem formulation}). Let $\O_1, \O_2$ and $\A_1, \A_2$ be the corresponding observation and action spaces. Let $H_{1,2}$ denote a space of functions from $\O_1$ to $\O_2$, and let $G_{2,1}$ denote a space of functions from $\A_2$ to $\A_1$. We will refer to a function $h \in H_{1,2}$ as an encoder and a function $g \in G_{2,1}$ as a decoder. 
\end{definition}

\begin{figure}[t]
    \centering
    \vspace{-5pt}
    \includegraphics[width=1.0\textwidth]{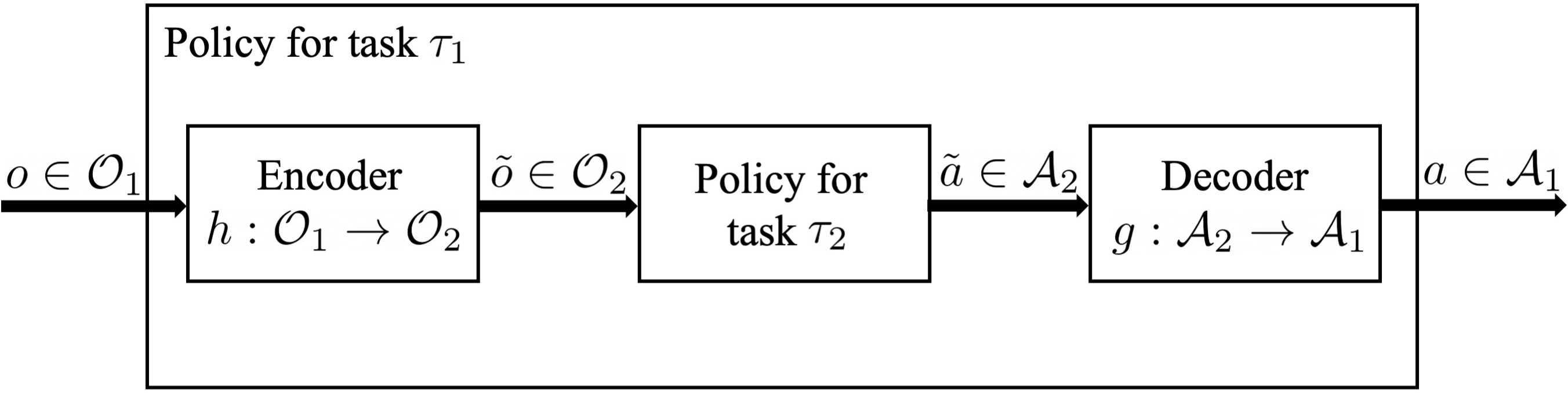}
    \vspace{-15pt}
    \caption{\smaller{The transformation of a policy for one task to another using \textit{encoders} and \textit{decoders}. An observation $o \in \O_1$ from task $\t_1$ is encoded to an observation $\tilde{o} \in \O_2$ for $\t_2$. The policy for $\t_2$ outputs an action $\tilde{a} \in \A_2$ which is decoded to an action $a \in \A_1$ for task $\t_1$. Together, the encoder, policy for $\t_2$, and decoder are a policy for $\t_1$.}}
    \label{fig:reduction}
    \vspace{-12pt}
\end{figure}

Intuitively, a task $\t_1$ reduces to $\t_2$ if we can utilize \textit{any} admissible policy for $\t_2$ and transform it via an encoder and decoder into a policy that solves $\t_1$; see Fig.~\ref{fig:reduction} for a visual representation of this transformation. Importantly, none of the elements which define a task ($\t := (\S, \A, \O, p, \sigma, r, p_0, R^\star)$) need to be shared between the two tasks. We formalize the notion of \emph{task reduction} below. 
% \begin{definition}[Task Reduction] Task $\t_1$ reduces to task $\t_2$ (written $\t_1 \tleq \t_2$) if there exists an encoder $h \in H_{1,2}$ and a decoder $g \in G_{2,1}$ such that: 
% \begin{equation}
%     g \circ \pi^\star_2 \circ h \in \Pi^\star_1
% \end{equation}
% for all admissible policies $\pi^\star_2 \in \Pi^\star_2$ for $\t_2$. 
% \label{def:task reduction}
% \end{definition}
\begin{definition}[Task Reduction] Task $\t_1$ reduces to task $\t_2$ (written $\t_1 \tleq \t_2$) if for all admissible policies $\pi^\star_2 \in \Pi^\star_2$, there exists an encoder $h \in H_{1,2}$ and a decoder $g \in G_{2,1}$ such that: 
\begin{equation}
    g \circ \pi^\star_2 \circ h \in \Pi^\star_1
\end{equation}
\label{def:task reduction}
\vspace{-5mm}
\end{definition}

This definition is illustrated in Fig.~\ref{fig:reduction}. This notion of reduction captures the relative complexity of tasks in terms of the complexity of online task execution. Thus, we compare the complexity of solutions rather than the complexity of computing solutions. It is important to note that the definition of reduction calls for the ability to transform \emph{any} admissible policy $\pi^\star_2 \in \Pi^\star_2$ into an admissible policy for $\tau_1$. 
% Thus $\Pi^\star_2$ is completely mapped into $\Pi^\star_1$ by functions from $H$ and $G$. 
We also note that task reductions are conditioned on the selection of $H_{1,2}$ and $G_{2,1}$.
Intuitively, this corresponds to the ``complexity'' of the reduction. For example, if $H_{1,2}$ and $G_{2,1}$ only include functions which can be evaluated in polynomial time, then the reduction $\t_1 \tleq \t_2$ is \emph{efficient} (analogous to Cook reductions). %  $\t_1$ is polynomial-time reducible to $\t_2$, analogous to Cook reductions. 
Additionally, note that a robot executing the policy $g \circ \pi^\star_2 \circ h \in \Pi^\star_1$ for task $\t_1$ over a time horizon of $T$ requires $T$ evaluations of $\pi^\star_2$. Thus, if $H_{1,2}$ and $G_{2,1}$ only include efficiently-computable functions and $\pi^\star_2$ is efficiently computable, then the \emph{online} execution of $\t_1$ is efficient. One can specify different classes of functions for $H_{1,2}$ and $G_{2,1}$ in order to capture different notions of efficiency (e.g., neural networks with bounded size). Next, we define \emph{equivalence} between two tasks. 
\begin{definition}[Task Equivalence] Tasks $\t_1$ and $\t_2$ are equivalent ($\t_1 \equiv \t_2$) if $\t_1 \tleq \t_2$ and $\t_2 \tleq \t_1$.
\label{def:task equality}
\end{definition}

Note that task reduction and equivalence are both binary relations (over $\T \times \T$). As we show below, these relations satisfy properties that are intuitively desirable for any definition of reduction (or equivalence) between robotic tasks. In order to state these properties, we first introduce a notion of function closure. %  which we will use will need when composing multiple task reductions.

\begin{definition}[Closure under Composition on $\T$] Let $g_1 \in G_{2,1}$, $g_2 \in G_{3,2}$, $h_1 \in H_{1,2}$, and $h_2 \in H_{2,3}$. We say that $G_{3,1}$ and $H_{1,3}$ are closed under composition if for any $\t_1,\t_2,\t_3 \in \T$, the following are true:
\begin{align}
    & g_1 \circ g_2 \in G_{3,1}, \ \forall \ g_1 \in G_{2,1} \  \forall \ g_2 \in G_{3,2}, \\ 
    & h_2 \circ h_1 \in H_{1,3}, \ \forall \ h_1 \in H_{1,2} \ \forall \ h_2 \in H_{2,3}.
\end{align} 
\label{def:reduction_composition} \vspace{-1em} \end{definition}
When $G_{\zeta,\xi}$ and $H_{\xi,\zeta}$ are the same for all tuples of tasks $(\t_\xi,\t_\zeta) \in \T^2$, this definition simplifies to closure under function composition for $G_{\zeta,\xi}$ and $H_{\xi,\zeta}$. % We now state and prove some basic but key properties of task reduction and equivalence. 

\begin{proposition}[Task Reduction is a Non-Strict Partial Ordering Relation] \label{prop:task reduction} 
% A relation has non-strict partial ordering if the following properties hold:
Suppose that $\forall \ (\t_\xi, \t_\zeta) \in \T^2$, $H_{\xi, \zeta}$ and $G_{\zeta, \xi}$ include the identity and are closed under composition on $\T$. Then, task reductions satisfy the following properties and thus define a non-strict partial ordering relation. 
\begin{customthm}{\ref{prop:task reduction}.a}. Reflexivity: $\t_1 \tleq \t_1$. \label{prop:reflexivity} \end{customthm} 
% \noindent Define $\t_1 \tless \t_2 := \t_1 \tleq \t_2 \wedge \neg(\t_1 \equiv \t_2)$.
\begin{customthm}{\ref{prop:task reduction}.b}. Antisymmetry: $\t_1 \tless \t_2 \implies \neg(\t_2 \tleq \t_1)$, where $\t_1 \tless \t_2$ is defined as $(\t_1 \tleq \t_2) \wedge \neg(\t_1 \equiv \t_2)$. \label{prop:antisymmetric} \end{customthm}
\begin{customthm}{\ref{prop:task reduction}.c}. Transitivity: $(\t_1 \tleq \t_2) \wedge (\t_2 \tleq \t_3) \implies \t_1 \tleq \t_3$.
\label{prop:transitivity} \end{customthm} 

\proof{A complete proof of these properties is provided in Appendix~\ref{ap:props}.}
\end{proposition}

Additionally, note that \textit{strict} task reduction $\t_1 \tless \t_2$ is a strict partial ordering relation. We provide a complete proof of this in Appendix~\ref{ap:props}. These properties formalize a series of intuitions about task reductions. Task reduction consists of transforming an admissible policy from one task to another. In the case of reflexivity, intuition suggest that if the tasks are the same, no transformation will be required for the policy to work. Thus, $G$ and $H$ must include the identity to allow for this. Antisymmetry establishes a clear sense of ``dominance'' between tasks in terms of complexity --- if a task $\t_1$ strictly reduces to another task $\t_2$, then $\t_2$ should not reduce to $\t_1$. Building on antisymmetry, we also require transitivity to make a clear chain of increasingly complex tasks. A series of tasks which reduce to each other cyclically (i.e. $\t_1 \tleq \t_2 \wedge \t_2 \tleq \t_3 \wedge \t_3 \tleq \t_1$) should only be possible if all the tasks are equivalent. The following proposition proves useful properties of task equivalence. 

\begin{proposition}[Task Equivalence is an Equivalence Relation] \label{prop:task equivalence} 
% A relation is an equivalence relation when the following properties hold:
Suppose that $\forall \ (\t_\xi, \t_\zeta) \in \T^2$, $H_{\xi, \zeta}$ and $G_{\zeta, \xi}$ include the identity and are closed under composition on $\T$. Then, task equivalence satisfies the following properties and thus defines an equivalence relation. 
\begin{customthm}{\ref{prop:task equivalence}.a}. Reflexivity: $\t_1 \equiv \t_1$. \label{prop:equiv_reflexive} \end{customthm}
\begin{customthm}{\ref{prop:task equivalence}.b}. Symmetry: $\t_1 \equiv \t_2 \implies \t_2 \equiv \t_1$. \label{prop:equiv_symmetric} \end{customthm}
\begin{customthm}{\ref{prop:task equivalence}.c}. Transitivity: $\t_1 \equiv \t_2 \wedge \t_2 \equiv \t_3 \implies \t_1 \equiv \t_3$. \label{prop:equiv_transitive}  \end{customthm}
\proof{A complete proof of these properties is provided in Appendix~\ref{ap:props}.} \end{proposition}
\section{Examples of Reductions}
\vspace{-5pt}
\label{sec:examples of reductions}
In this section, we provide two concrete examples of task reductions (Definition~\ref{def:task reduction}) and equivalence (Definition~\ref{def:task equality}). We present an efficient reduction between navigation tasks with differing goal locations in an environment which has $90$ degree rotational symmetry. We then show reduction between a set of tasks (i.e., an \textit{equivalence} class of tasks) for grasping with varying camera viewpoints. 

\subsection{Navigation to a Goal with Map Rotations}
Consider a robot which must navigate to a goal location and avoid obstacles. Suppose that the robot lives on an $[-n,n] \times [-n,n] \in \RR^2$ world and can move a distance $d$ in a cardinal direction $\{\mathrm{N}, \mathrm{E}, \mathrm{S}, \mathrm{W}\}$ by taking action $(d, \{0, 1, 2, 3\})$, where $0,1,2,3$ correspond to N,E,S,W. Additionally, for all tasks in this setting, assume that $m$ circular obstacles are randomly placed in the environment (while always allowing for a path to the goal for the robot). Suppose that the robot's observation at time $t$ corresponds to a complete map of the grid-world which consists of a list of locations including obstacle locations, the goal location, and the robot location. Let $o_\mathrm{E}^i = (x^i,y^i)$ correspond to the location of the $i^\mathrm{th}$ observation in $o_\mathrm{E}$. The robot is initialized in a random position and its goal is to navigate to the goal location. The robot receives a reward of 1 if it successfully navigates to the goal and a reward of 0 otherwise. 

Let task $\t_\mathrm{N}$ (``north'') have goal location at $(0, n)$ and $\t_\mathrm{E}$ (``east'') have goal location $(n, 0)$. An admissible policy $\pi_\mathrm{N}^\star$ on $\t_\mathrm{N}$ will always successfully navigate to the goal location $(0, n)$ (regardless of the initial state of the robot and the locations of the obstacles). 

\begin{proposition}[$\t_\mathrm{E} \tleq \t_\mathrm{N}$] \ Let $G_\mathrm{N,E}$ contain functions that can perform addition modulo 4 and $H_\mathrm{E,N}$ contain functions which can be evaluated in linear time (in the size of the observation). Then $\t_\mathrm{E} \tleq \t_\mathrm{N}$.
\proof{For any observation $o_\mathrm{E}$ from task $\t_\mathrm{E}$, let $h_\mathrm{EN}(o_\mathrm{E}^i) := (y^i, -x^i)\ \forall  \ i$. 
For any action $a_\mathrm{N} = (d, i)$ from $\pi^\star_\mathrm{N}$, let $g_\mathrm{NE}(d, i) := (d , i + 1 \mod 4)$.
Then we have that $g_{NE} \circ \pi^\star_\mathrm{N} \circ h_\mathrm{EN} \in \Pi^\star_\mathrm{E}, \  \forall \ \pi^\star_\mathrm{N} \in \Pi^\star_\mathrm{N}$. \qed }
\end{proposition}
Intuitively, $h_\mathrm{EN}$ rotates observations from $\t_\mathrm{E}$ so that they look exactly like a corresponding observation from $\t_\mathrm{N}$. Then, any admissible policy $\pi^\star_\mathrm{N}$ must find an admissible action (i.e., one which ultimately results in solving the task) based on the rotated observation. The decoder $g_{NE}$ then transforms the output action to the corresponding one required for task $\t_\mathrm{E}$. With analogous constructions, we can show that $\t_\mathrm{N} \equiv \t_\mathrm{E} \equiv \t_\mathrm{S} \equiv \t_\mathrm{E}$. % for the appropriate selection of $g$ and $h$. 

\subsection{Grasping Objects with Differing Camera Viewpoints} 
\label{subsec:graspingreduction}
Consider a robotic arm which must grasp one of a set of known objects using an RGB-D image. A randomly-selected object is placed with a random pose on a table (which is at height $z = 0$). %  from the robot can grasp. 
A task $\t \in \T$ will correspond to this grasping challenge when the RGB-D camera is placed at a particular viewpoint. Assume that the camera in each task is always pointed at the center of the table (i.e., $(0,0,0)$). % and always located $1$ unit away from the origin. 
Additionally, we will require that for each of the possible objects, the identity of the particular object and its pose are uniquely determinable %(up to symmetries of the object) 
from the camera. Further assume that at the beginning of the task, the robot arm does not occlude the camera's view of the object. We treat this as a single time-step task; the robot selects a grasp pose based on the camera observation and then controls $M$ motors in order to grasp the object. A successful grasp results in a reward of 1 and 0 otherwise. Let $\t_{(\theta,\phi)}$ correspond to the grasping task with the camera located at spherical coordinates $(1,\theta,\phi)$. 

\begin{proposition} [$\t_{(\theta_1,\phi_1)} \tleq \t_{(\theta_2,\phi_2)}$] For any $(\theta_1,\phi_1)$ and $(\theta_2,\phi_2)$, task  $\t_{(\theta_1,\phi_1)}$ reduces to task $\t_{(\theta_2,\phi_2)}$ if $(\theta_1, \phi_1)$ and $(\theta_2, \phi_2)$ are selected such that the cameras can view the entire table from either viewpoint.
\proof{ Let $g$ be the identity. Assume that the camera placed at $(\theta_1, \phi_1)$ can view the entire table. Let $h$ take as input a RGB-D image from viewpoint $(1, \theta_1,\phi_1)$ and output the same environmental setup from the perspective of a camera placed at $(1, \theta_2, \phi_2)$. Then $g \circ \pi^\star_{(\theta_2,\phi_2)} \circ h \in \Pi^\star_{(\theta_1,\phi_1)}$.
\qed}
\end{proposition}
Note that the function $h$ requires a model of the known objects and potentially a simulation of the environment in order to generate the image from a differing perspective. As such, the reduction with this encoder may not be efficient. A direct consequence of this proposition is that $\t_{(\theta_1,\phi_1)} \equiv \t_{(\theta_2,\phi_2)}$ if $(\theta_1, \phi_1)$ and $(\theta_2, \phi_2)$ are selected such that the cameras can view the entire table. Let $\Theta$ be the set of all $(\theta, \phi)$ values which reduce to each other using the reduction provided in the proposition. The corresponding set of tasks $\T_{\Theta} \subseteq \T$ defines a class of tasks which are all equivalently complex. 
\section{Relative Complexity}
\vspace{-5pt}
\label{sec:relative complexity}
% In this section, we tackle the second main goal of our work on comparing the complexity of robotic tasks. 
In general, it may be difficult to establish a reduction between two completely distinct tasks such as grasping an object and avoiding obstacles; indeed an arbitrary pair of tasks is unlikely to satisfy the notion of reduction introduced in Sec.~\ref{sec:reductions}. Additionally, our definition of reduction is a binary one and does not capture \emph{how} complex a particular task is relative to another. Motivated by these observations, we propose a definition of \emph{relative complexity} that captures the degree to which one task is more complex than another. This quantity can be thought of as a ``continuous'' or ``smoothed'' version of task reductions (in a precise sense, which we elucidate below). Note that we will omit the subscripts on $G$ and $H$ when it is clear which tasks they transform between.

\begin{definition}[Relative Complexity] The relative complexity of task $\t_1$ with respect to task $\t_2$ is
\label{def:relative complexity}
\begin{equation} \label{eq:relative complexity}
    C_{\t_1/\t_2}:=  \sup_{\pi^\star_2 \in \Pi^\star_2} \ \inf_{h\in H, g\in G} \bigg{[}1 - \frac{R_1(g\circ \pi^\star_2 \circ h)}{R^\star_1} \bigg{]},
\end{equation}
where we assume that rewards are nonnegative. We use the notation $C_{\t_1/\t_2}(H, G)$ when we want to highlight dependence on $H$ and $G$.
\end{definition}
% Note that the order of the min and max are important. As written, we assume that $H$ and $G$ are provided a policy $\pi^\star_2$ and are then allowed to select $h$ and $g$. 
Intuitively, if $\t_1 \tleq \t_2$, then this implies that $C_{\t_1 / \t_2} = 0$ (as we show formally below).
% if the relative complexity $C_{\t_1/\t_2} = 0$, this means that for any $\pi_2^\star \in \Pi_2^\star$ we can find an encoder $h$ and a decoder $g$ such that $R_1(g\circ \pi^\star_2 \circ h) = R^\star_1$. Hence, $\t_1$ is no more complex than $\t_2$ in this case. 
If $C_{\t_1/\t_2} > 0$, then an admissible policy $\pi_2^\star \in \Pi_2^\star$ may not be transformed into an admissible policy for $\t_1$ using encoders and decoders in $H$ and $G$. A key advantage of this definition is that we can compare \textit{any} two tasks and quantify the relative complexity of one with respect to another. As with our previous definitions, we prove a set of useful properties below. Importantly, Properties \ref{prop:relative_complexity 0 iff reduction} and \ref{prop:relative_complexity non0 iff noreduction} will establish a link between the relative complexity%$C_{\t_1/\t_2}$ 
 and task reduction. %  ($\t_1 \tleq \t_2$). 

\begin{proposition}[Properties of the Relative Complexity.] Relative Complexity satisfies the following properties: \label{prop:relative complexity}

\begin{customthm}{\ref{prop:relative complexity}.a}. Nonnegativity and boundedness: $C_{\t_1 / \t_2} \in  [0,1]$.
\label{prop:nonnegative} \end{customthm}
\begin{customthm}{\ref{prop:relative complexity}.b}. Monotonicity with respect to $H$ and $G$: If $H \subseteq H'$ and $G \subseteq G'$, then $C_{\t_1 / \t_2}(H', G') \leq C_{\t_1 / \t_2}(H, G)$.
\label{prop:monotonicity} \end{customthm}
 \noindent Assume that the supremum and infimum in Definition \ref{def:relative complexity} are attained by functions in $\Pi^\star_2, H, G$. Then:
\begin{customthm}{\ref{prop:relative complexity}.c}. Equivalence between reduction and $0$ relative complexity: $C_{\t_1 / \t_2} = 0 \iff \t_1 \tleq \t_2$.
% \edit{Reduction implies $0$ relative complexity: $\t_1 \tleq \t_2 \implies C_{\t_1 / \t_2} = 0.$} \vspace{-1em} 
\label{prop:relative_complexity 0 iff reduction} \end{customthm}

\begin{customthm}{\ref{prop:relative complexity}.d}.
% \edit{Positive relative complexity implies no reduction: $C_{\t_1 / \t_2} \in (0, 1]$\ $\implies \neg(\t_1 \tleq \t_2)$.} 
Equivalence between no reduction and positive relative complexity:  $C_{\t_1 / \t_2} \in (0,1] \iff \neg(\t_1 \tleq \t_2)$.
\label{prop:relative_complexity non0 iff noreduction} \end{customthm}
\vspace{-0.5em}
\proof{A complete proof of these properties is provided in Appendix~\ref{ap:props}.} \end{proposition}
\section{Algorithmic Approach}
\vspace{-5pt}
We can frame the optimization problem in \eqref{eq:relative complexity} as a two-player zero-sum game: an adversary chooses an admissible policy $\pi^\star_2 \in \Pi^\star_2$ which maximizes the relative complexity, and then the player chooses $g \in G$ and $h \in H$ such that the relative complexity is minimized. It may not be possible in general to find an optimal strategy to this game. However, we can still compute a meaningful estimate via approximate methods.

There are multiple ways to approximate optimal strategies for zero-sum games such as the one presented in \eqref{eq:relative complexity}; see e.g., \cite{fudenberg98}. One approach is to use \emph{best-response dynamics}, where the players update their strategies in rounds based on the \textit{best response} to the opponent's choice. We parameterize the policies, encoders, and decoders with neural networks. Thus, we aim to develop a method which uses gradient steps to approximate the solution to \eqref{eq:relative complexity}. A technique which exploits best-response dynamics is the update rule for training generative adversarial networks (GANs) \cite{goodfellow14}, where gradient steps update the players' strategies based on batches of data. The algorithm we present has a similar structure. 

The resulting approach has two steps and is presented in Algorithm~\ref{alg:relativecomplexity}. The first step is to update the policy $\pi_2$ to maximize the relative complexity. However, $\pi_2$ must also eventually succeed on task $\t_2$. Thus, the update for $\pi_2$ has two terms: one to train $\pi_2$ to succeed on $\t_2$ and one to train $g \circ \pi_2 \circ h$ to fail on $\t_1$. We scale the latter with the ``adversarial tuning parameter'' $\alpha$ which allows weighting of the terms relative to each other. The second step is to update the encoder and decoder to minimize the relative complexity. We present Algorithm~\ref{alg:relativecomplexity} with general loss functions $L_1, L_2$ on tasks $\t_1, \t_2$ to allow for specialization to reinforcement learning techniques (e.g., Q-learning). An example for $L_1$ and $L_2$ based on best-response dynamics is $L_2(\pi_2) := -R_2(\pi_2)$ and $L_1(g \circ \pi_2 \circ h) := 1 - R_1(g \circ \pi_2 \circ h)/ R_1^\star$ since these directly capture the objectives of each player.

In practice, when $\alpha$ is too low, the policy $\pi_2$ is not adversarial enough and may not prevent $g \circ \pi_2 \circ h$ from succeeding on task $\t_1$. When $\alpha$ is too large, the policy $\pi_2$ may not ever succeed on task $\t_2$. As such we increase $\alpha$ as much as possible while ensuring $\pi_2$ is admissible at the end of training in order to provide an estimate of \eqref{eq:relative complexity}. After training is complete, we have $\pi_2^\star$, $g$, and $h$ which we use to compute the approximate relative complexity $\tilde{C} = 1 - R_1(g\circ \pi^\star_2 \circ h)/R_1^\star$. 

\begin{algorithm}[t]
    \caption{Approximating Relative Complexity}
    \label{alg:relativecomplexity}
\begin{algorithmic}[1]
% \State \textbf{Input:} Threshold for success $R_1^\star, R_2^\star$ on $\t_1, \t_2$ respectively 
\State \textbf{Input:} Learning rates $\lambda_1, \lambda_2$, adversarial tuning parameter $\alpha$
\State \textbf{Input:} Function spaces  $H, G$
\State \textbf{Input:} Loss functions $L_1, L_2$ for $\t_1, \t_2$ respectively
\State \textbf{Output:} Approximate relative complexity $\tilde{C}_{\t_1/\t_2} \approx C_{\t_1/\t_2}$
\State \textbf{while} $\neg$(converged $\wedge$ $R_2(\pi_2) = R_2^\star$) \textbf{do}
\State \tab \textbf{Step 1: $\pi_2$ update} 
\State \tab $c_1 \leftarrow L_2(\pi_2) - \alpha L_1(g\circ \pi_2 \circ h)$
\State \tab $\pi_2 \leftarrow \pi_2 - \lambda_1\nabla_{\pi_2}c_1$
\State \tab \textbf{Step 2: encoder/decoder update} 
\State \tab $c_2 \leftarrow L_1(g\circ \pi_2 \circ h)$
\State \tab $[h,g] \leftarrow [h, g] - \lambda_2\nabla_{[h,g]}c_2$
\State \textbf{end while}
\State $\tilde{C}_{\t_1/\t_2} \leftarrow \bigg{[}1 - \frac{R_1(g\circ \pi^\star_2 \circ h)}{R_1^\star} \bigg{]}$ 
\end{algorithmic}
\end{algorithm}

\section{Examples}
\vspace{-5pt}
\label{sec:examples}
We implement Algorithm \ref{alg:relativecomplexity} on two reinforcement learning examples using Q-learning and Soft Actor-Critic (SAC). On the OpenAI Gym \cite{brockman16} Cartpole, we compare the complexity of balancing the pole upright relative to balancing it downwards. Using the Mujoco \cite{todorov12} 2D walker, we compare the complexity of walking at various speeds. We demonstrate that our estimates of relative complexity correspond to intuitive notions of complexity for these tasks. 

\subsection{Cartpole Balancing Task}
\label{sec:cartpole}

\textbf{Overview.} We use the OpenAI Gym \cite{brockman16} Cartpole environment to define two tasks: balancing the friction-less cart's pole against gravity $\t_\uparrow$ (at the unstable equilibrium) and balancing the cart's pole with gravity $\t_\downarrow$ (at the stable equilibrium). Equivalently, one can think of $\t_\uparrow$ and $\t_\downarrow$ being specified by the direction in which gravity acts ($-y$ and $+y$). The initial state of the system is randomized close to the equilibrium for each task. The policy receives the system's state vector as input and can apply forces on the cart using three actions: \{no force, push left, push right\}. A task runs for 200 time steps and the policy receives a reward of $1$ for each time step that the pole stays balanced (within $24^\circ$ of the equilibrium). Since we only switch the direction of gravity, the states which achieve a reward of $1$ are consistent between tasks. A reward of $0$ is given if the pole falls beyond $24^\circ$ of the equilibrium or the cart strays too far from the start position; the trial is then stopped. A policy for either task is admissible if it successfully balances the pole for the entire 200 time-step trial, i.e., $R_\downarrow^\star = R_\uparrow^\star = 200$. 

\begin{figure}[t]
    \centering
    \vspace{-17pt}
    \includegraphics[width=0.9\textwidth]{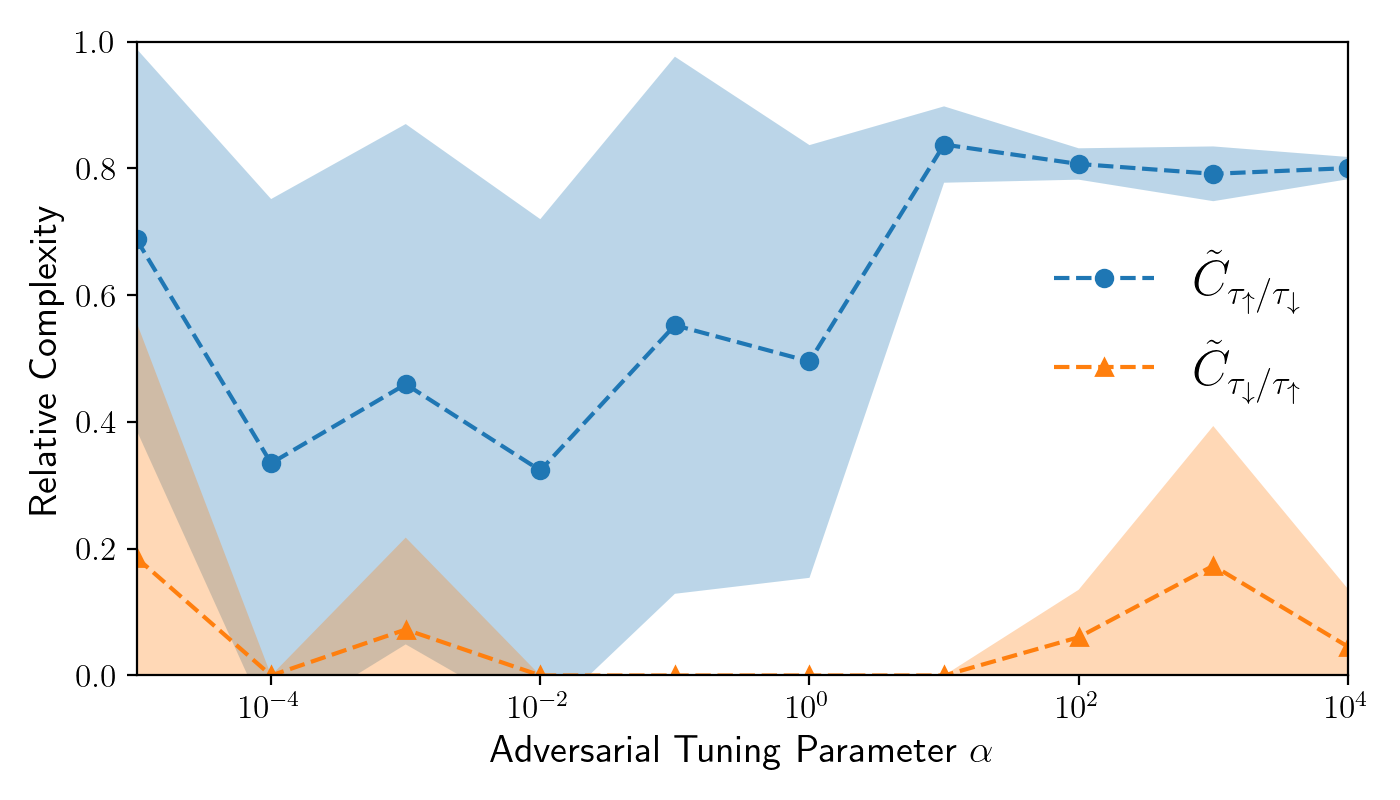}
    \vspace{-10pt}
    \caption{\smaller{The approximate relative complexity of task $\t_\uparrow$ with respect to $\t_\downarrow$ (circles) and $\t_\downarrow$ with respect to $\t_\uparrow$ (triangles) for varied adversarial tuning parameter $\alpha$. For large enough $\alpha$ ($\geq 10$), the relative complexity of $\t_\uparrow$ with respect to $\t_\downarrow$ is consistently close to $1$. In contrast, the relative complexity of $\t_\downarrow$ with respect to $\t_\uparrow$ is consistently close to $0$ regardless of the choice of $\alpha$. The plots suggest that $\t_\uparrow$ is \textit{more complex} than $\t_\downarrow$. We plot the mean and standard deviation (shaded region) across 5 seeds for each $\alpha$. \label{fig:cartpole_alpha}}}
    \vspace{-15pt}
\end{figure}

\textbf{Policy, Encoder, Decoder, and Training.} 
The policy, encoder, and decoder all consist of few-layer neural networks. We apply Algorithm \ref{alg:relativecomplexity} using Q-learning to approximate the relative complexities $C_{\t_\uparrow / \t_\downarrow}$ and $C_{\t_\downarrow / \t_\uparrow}$. % with $\tilde{C}_{\t_\uparrow / \t_\downarrow}$ and $\tilde{C}_{\t_\downarrow / \t_\uparrow}$ respectively. 
Note that we can directly apply Q-learning to Algorithm~\ref{alg:relativecomplexity} by letting the loss functions $L_1$ and $L_2$ correspond to a Q-learning loss (see Appendix~\ref{ap:exp_dets} for further details). In our experiments, we vary the adversarial tuning parameter to examine its effect and also vary the number of layers in encoder $h$ and decoder $g$ to approximate the relative complexity given varying complexity of $H$ and $G$. The result of training is an admissible policy $\pi^\star$, encoder $g$, and decoder $h$; these allow us to compute the estimated relative complexity $\tilde{C}$. 

\begin{figure}[t]
    \centering
    \vspace{-14pt}
    \includegraphics[width=0.8\textwidth]{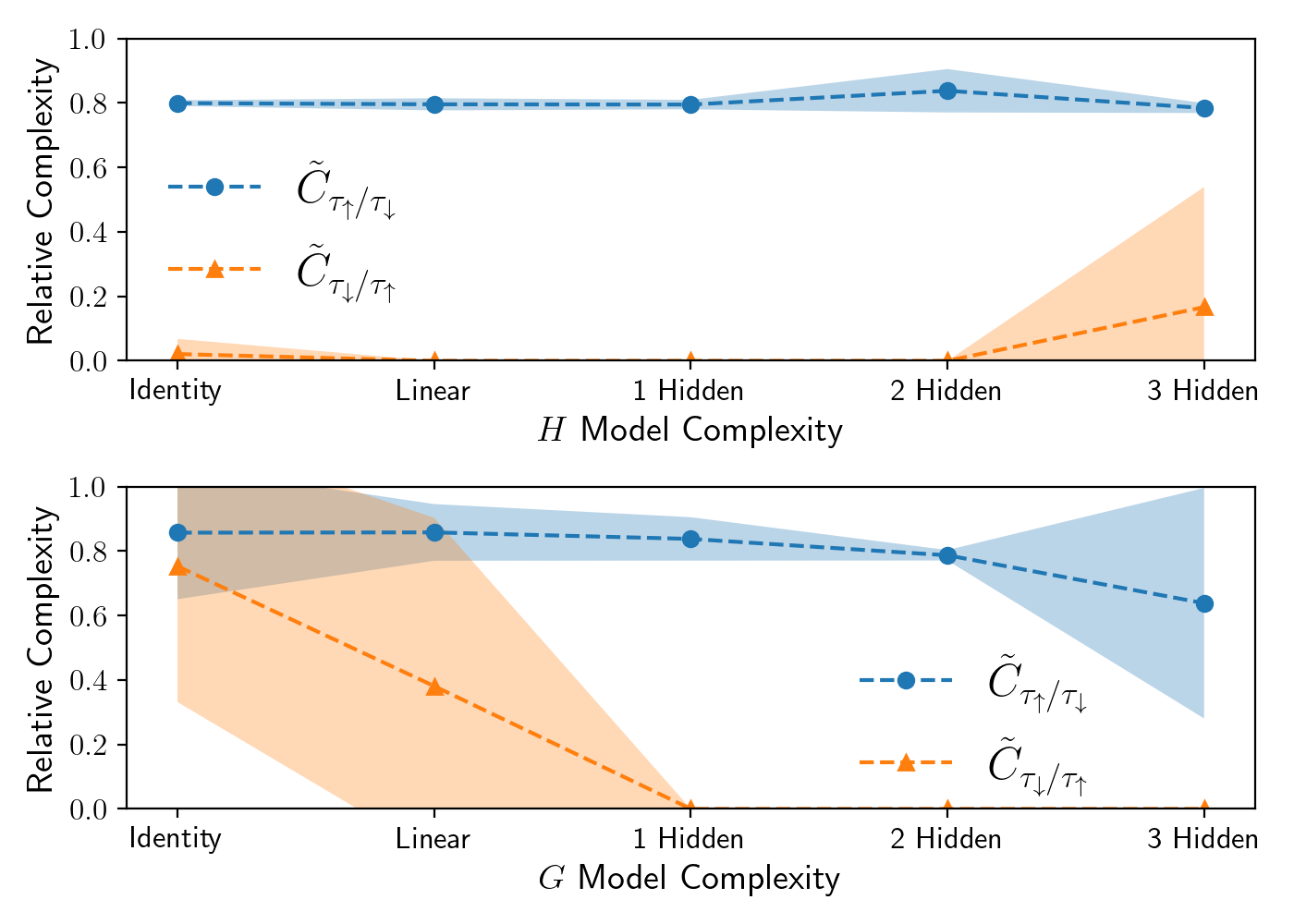}
    \vspace{-10pt}
    \caption{\smaller{The approximate relative complexity for varied $H$ and $G$ complexity. (Top) There is little or no change in the relative complexity for varied $H$. (Bottom) We see significant decrease in $C_{\t_\downarrow/\t_\uparrow}$ and a slight decrease in $C_{\t_\uparrow/\t_\downarrow}$ for increasing $G$ complexity. This suggest that reduction $\t_\downarrow \tleq \t_\uparrow$ is not possible when $G$ includes only linear neural networks. We plot the mean of 5 trials for each $\alpha$ value and the standard deviation is shaded. \label{fig:cartpole_models}}}
    \vspace{-10pt}
\end{figure}

\textbf{Results: Sensitivity to Adversarial Tuning Parameter.}
We compute the approximate relative complexity for a wide range of the adversarial tuning parameter $\alpha$: $[10^{-5}, 10^4]$, and plot the results in Fig.~\ref{fig:cartpole_alpha}. We choose $g$ to be a neural network with a single hidden layer and $h$ to have two hidden layers. When approximating the relative complexity of task $\t_\uparrow$ with respect to $\t_\downarrow$ we see that when $\alpha$ is large enough, the value settles at approximately $0.8$. This suggests that $\neg(\t_{\uparrow} \tleq \t_{\downarrow})$ by Property~\ref{prop:relative_complexity non0 iff noreduction}. Additionally, note that a decrease in $\alpha$ corresponds to decreased weight on minimizing $R_{\uparrow}(g \circ \pi_{\downarrow} \circ h)$. As such, when $\alpha$ is too small, $\pi_{\downarrow}$ cannot reliably ensure that $g \circ \pi_{\downarrow} \circ h$ fails on $\t_{\uparrow}$. In the case of approximating the relative complexity of task $\t_\downarrow$ with respect to $\t_\uparrow$, we see that for all values of $\alpha$, a low relative complexity is achieved. % \edit{The plot suggests that $C_{\t_\downarrow / \t_\uparrow} = 0$.}
The plot suggests that $\t_{\downarrow} \tleq \t_{\uparrow}$ by Property~\ref{prop:relative_complexity 0 iff reduction}. This suggests that the task of balancing at a stable equilibrium $\t_\downarrow$ is less complex than the task of balancing at an unstable equilibrium $\t_\uparrow$, which is consistent with our intuition.

\textbf{Results: Sensitivity to $H$ and $G$ Model Complexity.} 
We also approximate the relative complexity for different $H$ and $G$ with varying model complexity (achieved by varying the number of hidden layers). In all experiments, the neural network architecture for the (inner) policy $\pi$ is kept consistent. When varying the complexity of $H$, we choose $g$ to be a neural network with a single hidden layer; when varying $G$, we choose $h$ to be a neural network with two hidden layers. The results are plotted in Fig.~\ref{fig:cartpole_models}. We see that there is no significant change in the approximate relative complexity when $H$ increases in complexity. However, when the complexity of $G$ is increased, we see a clear decrease in $\tilde{C}_{\t_\downarrow/\t_\uparrow}$ indicating that (i) $\neg(\t_\downarrow \tleq \t_\uparrow)$ when $G$ only contains the identity or a single linear layer, and (ii) $\t_\downarrow \tleq \t_\uparrow$ for more complex $G$. We see a slight decrease in $\tilde{C}_{\t_\uparrow/\t_\downarrow}$ for increasing complexity of $G$. By Property~\ref{prop:monotonicity}, the relative complexity is monotonic with respect to $H$ and $G$; specifically, increasing the complexity of $H$ and $G$ should result in a monotonic decrease in the relative complexity. This is consistent with the results in Fig.~\ref{fig:cartpole_models}.

\subsection{Mujoco 2D Walker at varied speeds}
\label{sec:walker}
\textbf{Overview.}
Using the Mujoco \cite{todorov12} Walker2D-v2 environment in OpenAI Gym \cite{brockman16}, we create a set of tasks $\t_v$ with the goal of walking at a particular speed $v$ (from 0.6 $m$/$s$  to 1.4 $m$/$s$). Thus, the maximum reward on $\t_v$ is achieved when the robot travels at $v$. The policy receives a 17-dimensional observation vector (corresponding to joint angles and velocities) and outputs a 6-dimensional action vector (corresponding to joint torques) to control the 2D walking robot. The task runs for at most 1000 time steps. The reward at time $t$ is $1 - |v_t - v|$, where $v_t$ is the speed at the current time step. Note that to help with training, we also provide a reward for staying upright and a reward which penalizes large policy outputs to help with training. When evaluating $\tilde{C}$, we lower-bound the reward by $0$ to ensure that the reward is non-negative. If the robot falls over, a reward of $0$ is given and the trial stops. Since the challenge of maintaining a particular speed may differ between tasks, the threshold for success may vary. Thus, to find $R_v^\star$, we first train an individual policy to travel at speed $v$. We then let $R_v^\star = 0.95 \ \times$   (individual policy reward on $\t_v$).

\begin{figure}[t]
    \centering
    \vspace{-10pt}
    \includegraphics[width=0.8\textwidth]{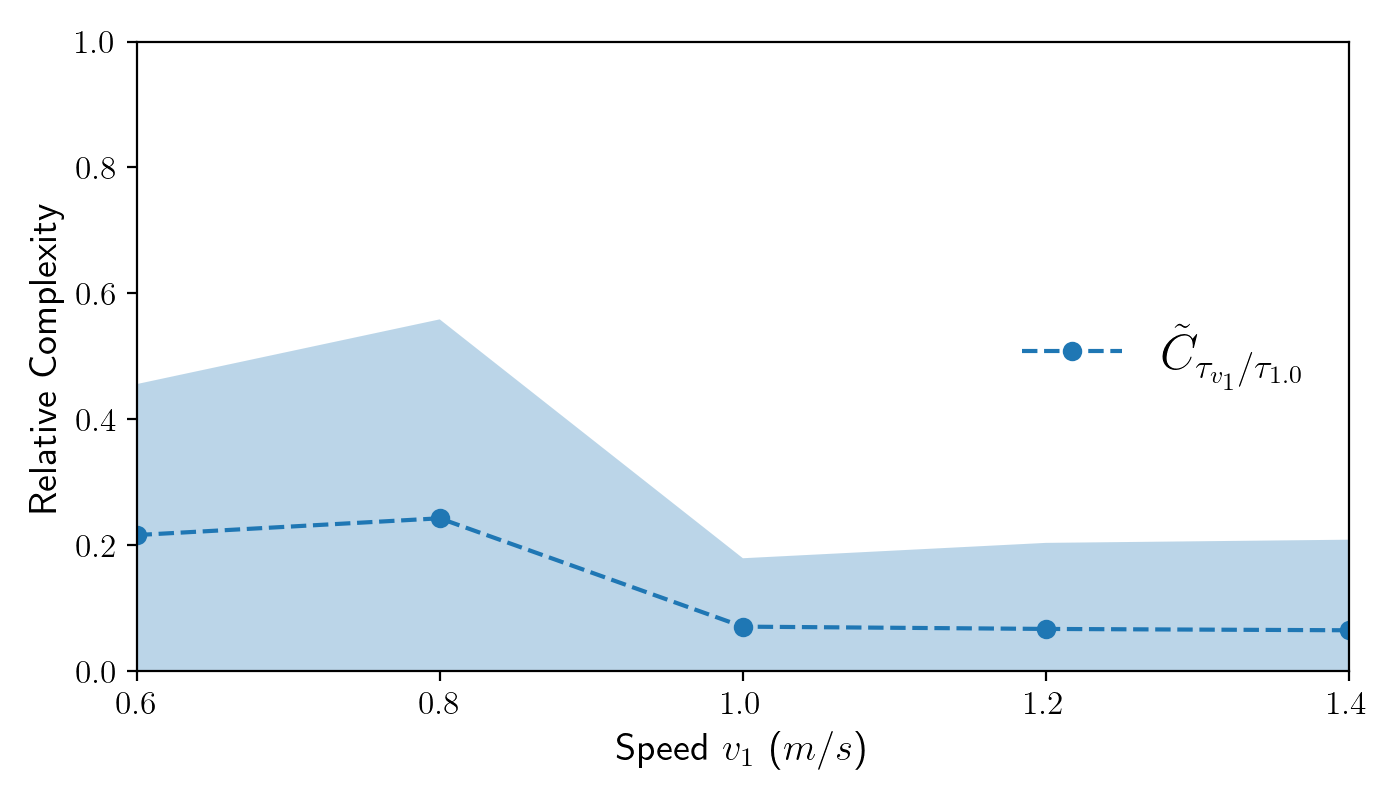}
    \vspace{-10pt}
    \caption{\smaller{Reduction from the task of walking at speed $v_1$ ($\t_{v_1}$) to walking at $1.0$ $m$/$s$ ($\t_{1.0}$). The estimated relative complexity $\tilde{C}_{\t_{v_1}/\t_{1.0}}$ may be larger when $v_1 < 1.0$ $m$/$s$ as compared with $v_1 \geq 1.0$ $m$/$s$. Thus, it is possible that a policy for walking at $1.0$ $m$/$s$ cannot directly be transformed to a policy for walking at $0.6$ or $0.8$ $m$/$s$. % The plot suggests that $\t_{v_1} \tleq \t_{1.0}$ for $v_1 \in \{1.0, 1.2, 1.4\}$ $m$/$s$ but $\neg(\t_{v_1} \tleq \t_{1.0})$ for $v_1 \in \{0.6, 0.8\}$. 
    We plot the mean over 15 seeds for each $v_1$; the standard deviation is shaded.} 
    \label{fig:speed_walker}}
    \vspace{-10pt}
\end{figure}

\textbf{Policy, Encoder, Decoder and Training.}
As in Sec.~\ref{sec:cartpole}, the policy, encoder, and decoder all consist of neural networks with 3 hidden layers. However, since the action space is continuous and $6$-dimensional, we use soft actor-critic (SAC) \cite{haarnoja18}. This requires a modification to Algorithm~\ref{alg:relativecomplexity} for training critics for $\t_{v_1}$ and $\t_{v_2}$ (see Appendix~\ref{ap:exp_dets} for the algorithm and additional experimental details). The result of training is an admissible policy $\pi^*_2$ and the encoder/decoder $h$/$g$; these are used to calculate the approximate relative complexity $\tilde{C}$. 

\textbf{Results: Relative Complexity for Increasing Speed.}
We compute the approximate relative complexity of the task of walking at speeds $v_1 \in [0.6, 1.4]$ $m$/$s$ with respect to the task of walking at $1.0$ $m$/$s$ ($\t_{1.0}$). We sweep through a wide range of the adversarial tuning parameter $\alpha = [10^{-6}, 10]$, and for each $\tilde{C}$ computation, choose the largest $\alpha$ which results in an admissible policy on task $\t_{1.0}$.
The results are presented in Fig.~\ref{fig:speed_walker}. 
% \edit{Importantly, note that these results are conditioned on the choice of $H$, $G$, and are kept consistent for all relative complexity computations.} 
The choice of $H$ and $G$ is kept consistent for all relative complexity computations. We see that $\tilde{C}_{\t_{v_1}/\t_{1.0}}$ may be larger when $v_1 < 1.0$ $m$/$s$ as compared with $v_1 \geq 1.0$ $m$/$s$. Importantly, these results are conditioned on the choice of $H$, and $G$. The plot suggests that there exists an admissible policy for walking at $1.0$ $m$/$s$ which cannot be directly transformed to walk at $0.6$ or $0.8$ $m$/$s$ but no such policy exists which prevents a transformation to a policy for walking faster. % Additionally these results indicate that $\t_{v_1} \tleq \t_{1.0}$ for $v_1 \in \{1.0, 1.2, 1.4\}$ $m$/$s$ but $\neg(\t_{v_1} \tleq \t_{1.0})$ for $v_1 \in \{0.6, 0.8\}$.

\section{Conclusion}
\vspace{-5pt}

We have presented a framework for comparing the complexity of robotic tasks. In order to achieve this, we defined a notion of reduction between two tasks that captures the ability of a robot to solve one task given a policy for another. We also presented a measure of relative complexity that quantifies how complex one task is relative to another. Our theoretical results establish basic properties satisfied by these notions and also establish the relationship between reductions and relative complexity. We presented a practical algorithm for estimating the relative complexity between tasks and demonstrated this using reinforcement learning tasks. Our results demonstrate consistency with intuitive notions of hardness for these tasks and empirical correspondence to theoretical properties. 

The work presented here opens up a number of exciting directions for future research. We begin by discussing some of the limitations of this framework as motivation. One such limitation is the slow convergence of the computation for approximating the relative complexity (which required longer to converge than learning a policy for either individual task). We also need admissible policies for both $\t_1$ and $\t_2$ to compute the relative complexity. Additionally, the results we present are conditioned on the choice of encoder $H$ and decoder $G$ spaces. The value and interpretability of the relative complexity measure thus depend on how easily one can interpret the expressiveness of $H$ and $G$.

On the theoretical front, it would be interesting to devise more general notions of reductions than the one presented here, e.g., allowing the policy from one task to be used as an oracle to solve another task in a more general manner (instead of the specific encoder-policy-decoder architecture we utilize). Another research direction is to use our notion of reduction to establish equivalences between broad classes of tasks (similar to complexity classes P and NP in computational complexity). It may also be of practical interest to define relative complexity in terms of different metrics such as data efficiency or difficulty of finding a policy instead of the notion of complexity we use, which is based on the complexity of online execution of policies.

On the algorithmic front, it would be of practical interest to develop different techniques for estimating the relative complexity. One could potentially use other algorithms (beyond best-response dynamics) for approximately solving games \cite{fudenberg98}. Finally, a particularly exciting direction would be to draw inspiration from \emph{cryptography} and turn statements about the complexity of certain tasks into statements about \emph{robustness} for a robot. This could potentially be achieved by establishing the hardness of an \emph{adversary's} task of foiling our robot.

\subsection*{Acknowledgements}
% \vspace{-3pt}
\footnotesize{The authors are grateful to the anonymous reviewers for their helpful feedback and suggestions on this work. Funding: NSF CAREER Award [\#2044149] and Office of Naval Research [N00014-21-1-2803, N00014-18-1-2873].}
\bibliographystyle{splncs04}
% \vspace{-5pt}
\begin{footnotesize}
\bibliography{citations.bib}

\begin{thebibliography}{10}
\providecommand{\url}[1]{\texttt{#1}}
\providecommand{\urlprefix}{URL }
\providecommand{\doi}[1]{https://doi.org/#1}

\bibitem{achille19}
Achille, A., Mbeng, G., Soatto, S.: Dynamics and reachability of learning
  tasks. arXiv:1810.02440  (2019)

\bibitem{achille20}
Achille, A., Paolini, G., Mbeng, G., Soatto, S.: The information complexity of
  learning tasks, their structure and their distance. Information and
  Inference: A Journal of the IMA  \textbf{10}(1),  51--72 (2021)

\bibitem{Ahmadi13}
Ahmadi, A.A., Majumdar, A., Tedrake, R.: Complexity of ten decision problems in
  continuous time dynamical systems. Proceedings of the American Control
  Conference (ACC) pp. 6376--6381 (2013)

\bibitem{Arora09}
Arora, S., Barak, B.: Computational Complexity: A Modern Approach. Cambridge
  University Press, New York, NY (2009)

\bibitem{Blondel00}
Blondel, V., Tsitsiklis, J.: A survey of computational complexity results in
  systems and control. Automatica  \textbf{36}(9),  1249--1274 (2000)

\bibitem{Borie09}
Borie, R., Tovey, C., Koenig, S.: Algorithms and complexity results for
  pursuit-evasion problems. International Joint Conference on Artificial
  Intelligence (IJCAI)  \textbf{9},  59--66 (2009)

\bibitem{Borie11}
Borie, R., Tovey, C., Koenig, S.: Algorithms and complexity results for
  graph-based pursuit evasion. Autonomous Robots  \textbf{31}(4),  317--332
  (2011)

\bibitem{brockman16}
Brockman, G., Cheung, V., Pettersson, L., Schneider, J., Schulman, J., Tang,
  J., Zaremba, W.: Openai gym. arXiv:1606.01540  (2016)

\bibitem{Canny88}
Canny, J.: The Complexity of Robot Motion Planning. MIT press, Cambridge, MA
  (1988)

\bibitem{Chang19}
Chang, M.B., Gupta, A., Levine, S., Griffiths, T.L.: Automatically composing
  representation transformations as a means for generalization. Proceedings of
  the International Conference on Learning Representations  (2019)

\bibitem{Culberson97}
Culberson, J.: Sokoban is {PSPACE}-complete. Tech. Rep. TR 97-02, University of
  Alberta, Edmonton, Alberta, Canada (1997)

\bibitem{Donald95}
Donald, B.R.: On information invariants in robotics. Artificial Intelligence
  \textbf{72}(1),  217--304 (1995)

\bibitem{Erdmann95}
Erdmann, M.: Understanding action and sensing by designing action-based
  sensors. The International Journal of Robotics Research (IJRR)
  \textbf{14}(5),  483--509 (1995)

\bibitem{fudenberg98}
Fudenberg, D., Drew, F., Levine, D.K.: The theory of learning in games, vol.~2.
  MIT press, Cambridge, MA (1998)

\bibitem{goodfellow14}
Goodfellow, I., Pouget-Abadie, J., Mirza, M., Xu, B., Warde-Farley, D., Ozair,
  S., Courville, A., Bengio, Y.: Generative adversarial nets. Advances in
  Neural Information Processing Systems  \textbf{27} (2014)

\bibitem{haarnoja18}
Haarnoja, T., Zhou, A., Abbeel, P., Levine, S.: Soft actor-critic: Off-policy
  maximum entropy deep reinforcement learning with a stochastic actor.
  Proceedings of the International Conference on Machine Learning pp.
  1861--1870 (2018)

\bibitem{Han17}
Han, S., Stiffler, N., Krontiris, A., Bekris, K., Yu, J.: High-quality tabletop
  rearrangement with overhand grasps: Hardness results and fast methods.
  Proceedings of Robotics: Science and Systems (RSS)  (2017)

\bibitem{Hauser14}
Hauser, K.: The minimum constraint removal problem with three robotics
  applications. The International Journal of Robotics Research (IJRR)
  \textbf{33}(1),  5--17 (2014)

\bibitem{Hopcroft84a}
Hopcroft, J., Joseph, D., Whitesides, S.: Movement problems for 2-dimensional
  linkages. SIAM Journal on Computing (SICOMP)  \textbf{13}(3),  610--629
  (1984)

\bibitem{Hopcroft84}
Hopcroft, J., Schwartz, J., Sharir, M.: On the complexity of motion planning
  for multiple independent objects; {PSPACE}-hardness of the {W}arehouseman's
  {P}roblem. The International Journal of Robotics Research (IJRR)
  \textbf{3}(4),  76--88 (1984)

\bibitem{Joseph85}
Joseph, D., Plantings, W.H.: On the complexity of reachability and motion
  planning questions. Proceedings of the Symposium on Computational Geometry
  pp. 62--66 (1985)

\bibitem{Lavalle06}
{LaValle}, S.M.: Planning Algorithms. Cambridge University Press, Cambridge, MA
  (2006)

\bibitem{Lavalle12}
LaValle, S.M.: Sensing and Filtering: A Fresh Perspective Based on Preimages
  and Information Spaces. Publishers Inc., Hanover, MA (2012)

\bibitem{Li08}
Li, M., Vit{\'a}nyi, P., et~al.: An introduction to {K}olmogorov complexity and
  its applications, vol.~3. Springer, New York, NY (2008)

\bibitem{Li21}
Li, Y., Wu, Y., Xu, H., Wang, X., Wu, Y.: Solving compositional reinforcement
  learning problems via task reduction. Proceedings of the International
  Conference on Learning Representations  (2021)

\bibitem{Murrieta08}
Murrieta-Cid, R., Monroy, R., Hutchinson, S., Laumond, J.P.: A {C}omplexity
  result for the pursuit-evasion game of maintaining visibility of a moving
  evader. Proceedings of the IEEE International Conference on Robotics and
  Automation (ICRA) pp. 2657--2664 (2008)

\bibitem{OKane08}
O'Kane, J.M., LaValle, S.M.: Comparing the {Power} of {Robots}. The
  International Journal of Robotics Research (IJRR)  \textbf{27}(1),  5--23
  (2008)

\bibitem{Reif79}
Reif, J.: Complexity of the mover's problem and generalizations. Symposium on
  Foundations of Computer Science pp. 421--427 (1979)

\bibitem{Saberifar19}
Saberifar, F.Z., Ghasemlou, S., Shell, D.A., O’Kane, J.M.: Toward a
  language-theoretic foundation for planning and filtering. The International
  Journal of Robotics Research (IJRR)  \textbf{38}(2-3),  236--259 (2019)

\bibitem{OKane20}
Shell, D.A., O’Kane, J.M.: Reality as a simulation of reality: robot
  illusions, fundamental limits, and a physical demonstration. Proceedings of
  the IEEE International Conference on Robotics and Automation (ICRA) pp.
  10327--10334 (2020)

\bibitem{Sipser96}
Sipser, M.: Introduction to the Theory of Computation. Cengage Learning,
  Boston, MA (2013)

\bibitem{Solovey16}
Solovey, K., Halperin, D.: On the hardness of unlabeled multi-robot motion
  planning. The International Journal of Robotics Research (IJRR)
  \textbf{35}(14),  1750--1759 (2016)

\bibitem{todorov12}
Todorov, E., Erez, T., Tassa, Y.: Mujoco: A physics engine for model-based
  control. Proceedings of the IEEE/RSJ International Conference on Intelligent
  Robots and Systems (IROS) pp. 5026--5033 (2012)

\bibitem{Tran19}
Tran, A.T., Nguyen, C.V., Hassner, T.: Transferability and hardness of
  supervised classification tasks. Proceedings of the IEEE/CVF International
  Conference on Computer Vision pp. 1395--1405 (2019)

\end{thebibliography}
\end{footnotesize}

\newpage
\appendix
\section*{Appendix}
\section{Proof of Properties}
\label{ap:props}

\begin{customt}{\ref{prop:task reduction}}{\normalfont\textbf{(Task Reduction is a Non-Strict Partial Ordering Relation).}}
% A relation has non-strict partial ordering if the following properties hold:
Suppose that $\forall \ (\t_\xi, \t_\zeta) \in \T^2$, $H_{\xi, \zeta}$ and $G_{\zeta, \xi}$ include the identity and are closed under composition on $\T$. Then, task reductions satisfy the following properties and thus define a non-strict partial ordering relation. 
\begin{customthm}{\ref{prop:task reduction}.a}. Reflexivity: $\t_1 \tleq \t_1$. \end{customthm} 
% \noindent Define $\t_1 \tless \t_2 := \t_1 \tleq \t_2 \wedge \neg(\t_1 \equiv \t_2)$.
\begin{customthm}{\ref{prop:task reduction}.b}. Antisymmetry: $\t_1 \tless \t_2 \implies \neg(\t_2 \tleq \t_1)$, where $\t_1 \tless \t_2$ is defined as $(\t_1 \tleq \t_2) \wedge \neg(\t_1 \equiv \t_2)$. \end{customthm}
\begin{customthm}{\ref{prop:task reduction}.c}. Transitivity: $(\t_1 \tleq \t_2) \wedge (\t_2 \tleq \t_3) \implies \t_1 \tleq \t_3$.
\end{customthm} 

\proof{\textbf{Property \ref{prop:reflexivity}:} $\t_1 \tleq \t_1 \implies \exists \ g \in G_{1,1}$, $h \in H_{1,1}$ such that 
\begin{equation}
    g \circ \pi^\star_1 \circ h \in \Pi^\star_1.
\end{equation}
If $g$ and $h$ are the identity function, then $g \circ \pi^\star_1 \circ h = \pi^\star_1 \ \forall \ \pi^\star_1 \in \Pi^\star_1$. Thus, $\t_1 \tleq \t_1$ when $G_{1,1}$ and $H_{1,1}$ include their respective identity functions. 

\textbf{Property \ref{prop:antisymmetric}:}
Suppose $\t_1 \tless \t_2$ and thus $(\t_1 \tleq \t_2) \wedge \neg(\t_1 \equiv \t_2)$. Note that $\neg(\t_1 \equiv \t_2) \implies \neg \big{(}(\t_1 \tleq \t_2) \wedge (\t_2 \tleq \t_1) \big{)} \implies \neg(\t_1 \tleq \t_2) \vee \neg(\t_2 \tleq \t_1)$. We assumed $(\t_1 \tleq \t_2)$, so we must have that $\neg(\t_2 \tleq \t_1)$. % Thus, task reductions are antisymmetric.

\textbf{Property \ref{prop:transitivity}:} Suppose $\t_1 \tleq \t_2$ and $\t_2 \tleq \t_3$. By Definition \ref{def:task reduction}, $\exists \ g_1 \in G_{2,1}$, $g_2 \in G_{3,2}$, $h_1 \in H_{1,2}$, and $h_2 \in H_{2,3}$ such that $g_1 \circ \pi^\star_2 \circ h_1 \in \Pi^\star_1 \ \forall \ \pi^\star_2 \in \Pi^\star_2$ and $g_2 \circ \pi^\star_3 \circ h_2 \in \Pi^\star_2 \ \forall \ \pi^\star_3 \in \Pi^\star_3$. Consider
\begin{equation}
    \underbrace{g_1 \circ \overbrace{g_2 \circ \pi^\star_3 \circ h_2}^{\in \Pi^\star_2} \circ h_1}_{\in \Pi^\star_1}
\end{equation}
for all $\pi^\star_3 \in \Pi^\star_3$.
Let $g_3 := g_1 \circ g_2$ and $h_3 := h_2 \circ h_1$ so that $g_3 \circ \pi^\star_3 \circ h_3 \in \Pi^\star_1$ for all $\pi^\star_3 \in \Pi^\star_3$.
If $G_{3,1}$ and $H_{1,3}$ are closed under composition on $\T$, then $g_3 \in G_{3,1}$ and $h_3 \in H_{1,3}$ and $\t_1 \tleq \t_3$. Thus, task reductions are transitive if $G_{3,1}$ and $H_{1,3}$ are closed under composition on $\T$. \qed
} \end{customt}

\begin{proposition}[Strict Task Reduction is a Strict Partial Ordering Relation] \label{prop:strict task reduction} Suppose that $\forall \ (\t_\xi, \t_\zeta) \in \T^2$, $H_{\xi, \zeta}$ and $G_{\zeta, \xi}$ include the identity and are closed under composition on $\T$. Then, strict task reductions satisfy the following properties and thus define a strict partial ordering relation.
\begin{customthm}{\ref{prop:strict task reduction}.a}. Irreflexivity: $\neg(\t_1 \tless \t_1)$. \label{prop:irreflexivity} \end{customthm} 
\begin{customthm}{\ref{prop:strict task reduction}.b}. Asymmetry: $\t_1 \tless \t_2 \implies \neg(\t_2 \tless \t_1)$. \label{prop:asymmetry} \end{customthm}
\begin{customthm}{\ref{prop:strict task reduction}.c}. Transitivity: $\t_1 \tless \t_2 \wedge \t_2 \tless \t_3 \implies \t_1 \tless \t_3$.
\label{prop:strict transitivity} \end{customthm} 

\proof{\textbf{Property \ref{prop:irreflexivity}:} Suppose $\t_1 \tless \t_1 \implies \t_1 \tleq \t_1 \wedge \neg(\t_1 \equiv \t_1)$. $\t_1 \equiv \t_1$ by Property \ref{prop:equiv_symmetric}. $\Rightarrow\!\Leftarrow \implies \neg(\t_1 \tless \t_1)$ when $H_{1,1}$ and $G_{1,1}$ include their respective identity functions. 

\textbf{Property \ref{prop:asymmetry}:} $\t_1 \tless \t_2 \implies \neg(\t_2 \tleq \t_1)$ by Property \ref{prop:antisymmetric}, since $\t_1 \tless \t_2 \implies \neg(\t_1 \equiv \t_2)$. $\neg(\t_2 \tleq \t_1) \iff \neg(\t_2 \tleq \t_1) \vee \t_2 \equiv \t_1 \implies \neg\big(\t_2 \tleq \t_1 \wedge \neg(\t_2 \equiv \t_1)\big) \implies \neg(\t_2 \tless \t_1).$

\textbf{Property \ref{prop:strict transitivity}:} $\t_1 \tless \t_2 \wedge \t_2 \tless \t_3 \implies \t_1 \tleq \t_2 \wedge \t_2 \tleq \t_3 \wedge \neg(\t_1 \equiv \t_2) \wedge \neg(\t_2 \equiv \t_3) \implies \t_1 \tleq \t_3 \wedge \neg(\t_1 \equiv \t_3)$ by Properties \ref{prop:transitivity} and \ref{prop:equiv_transitive}. $\implies \t_1 \tless \t_3$ when $H_{1,3}$ and $G_{3,1}$ are closed under composition on $\T$.
\qed
} \end{proposition}

\begin{customt}{\ref{prop:task equivalence}}{\normalfont\textbf{(Task Equivalence is an Equivalence Relation).}}
Suppose that $\forall \ (\t_\xi, \t_\zeta) \in \T^2$, $H_{\xi, \zeta}$ and $G_{\zeta, \xi}$ include the identity and are closed under composition on $\T$. Then, task equivalence satisfies the following properties and thus defines an equivalence relation. 
\begin{customthm}{\ref{prop:task equivalence}.a}. Reflexivity: $\t_1 \equiv \t_1$. \end{customthm}
\begin{customthm}{\ref{prop:task equivalence}.b}. Symmetry: $\t_1 \equiv \t_2 \implies \t_2 \equiv \t_1$. \end{customthm}
\begin{customthm}{\ref{prop:task equivalence}.c}. Transitivity: $\t_1 \equiv \t_2 \wedge \t_2 \equiv \t_3 \implies \t_1 \equiv \t_3$. \end{customthm}

\proof{
\textbf{Property \ref{prop:equiv_reflexive}:} 
$\t_1 \equiv \t_1 \implies \t_1 \tleq \t_1$ by Property \ref{prop:reflexivity} when $G_{1,1}$ and $H_{1,1}$ include the identity. Thus, task equivalence is reflexive if $G_{1,1}$ and $H_{1,1}$ include the identity. 

\textbf{Property \ref{prop:equiv_symmetric}:}
$\t_1 \equiv \t_2 \implies (\t_1 \tleq \t_2) \wedge (\t_2 \tleq \t_1)$ by Definition \ref{def:task equality} $\implies (\t_2 \tleq \t_1) \wedge (\t_1 \tleq \t_2)$ $\implies \t_2 \equiv \t_1$.

\textbf{Property \ref{prop:equiv_transitive}:} 
$(\t_1 \equiv \t_2) \wedge (\t_2 \equiv \t_3) \implies (\t_1 \tleq \t_2) \wedge (\t_2 \tleq \t_3) \wedge (\t_3 \tleq \t_2) \wedge (\t_2 \tleq \t_1)$ by Definition \ref{def:task equality}. $(\t_3 \tleq \t_2) \wedge (\t_2 \tleq \t_1) \implies (\t_3 \tleq \t_1)$ by Property \ref{prop:transitivity} when $G_{3,1}$ and $H_{1,3}$ are closed under composition on $\T$. Similarly, $(\t_1 \tleq \t_2) \wedge (\t_2 \tleq \t_3) \implies (\t_1 \tleq \t_3)$. Thus $(\t_1 \tleq \t_3) \wedge (\t_3 \tleq \t_1) \implies \t_1 \equiv \t_3$. Thus task equivalence is transitive if $G_{3,1}$ and $H_{1,3}$ are closed under composition on $\T$.  \qed
}
\end{customt}

\begin{customt}{\ref{prop:relative complexity}}{\normalfont\textbf{(Properties of the Relative Complexity).}} Relative Complexity satisfies the following properties:

\begin{customthm}{\ref{prop:relative complexity}.a}. Nonnegativity and boundedness: $C_{\t_1 / \t_2} \in  [0,1]$. \end{customthm}
\begin{customthm}{\ref{prop:relative complexity}.b}. Monotonicity with respect to $H$ and $G$: If $H \subseteq H'$ and $G \subseteq G'$, then $C_{\t_1 / \t_2}(H', G') \tleq C_{\t_1 / \t_2}(H, G)$. \end{customthm}
\noindent Assume that the supremum and infimum in Definition \ref{def:relative complexity} are attained by functions in $\Pi^\star_2, H, G$. Then:
\begin{customthm}{\ref{prop:relative complexity}.c}. 
Equivalence between reduction and $0$ relative complexity: $C_{\t_1 / \t_2} = 0 \iff \t_1 \tleq \t_2$.
% \edit{Reduction implies $0$ relative complexity: $\t_1 \tleq \t_2 \implies C_{\t_1 / \t_2} = 0.$}
\end{customthm}
\begin{customthm}{\ref{prop:relative complexity}.d}. 
% \edit{Positive relative complexity implies no reduction: $C_{\t_1 / \t_2} \in (0,1] \implies \neg(\t_1 \tleq \t_2)$}
Equivalence between no reduction and positive relative complexity:  $C_{\t_1 / \t_2} \in (0,1] \iff \neg(\t_1 \tleq \t_2)$.
\end{customthm}

\proof{
\textbf{Property \ref{prop:nonnegative}:} $R_1(g \circ \pi^\star_2 \circ h) \in [0, R^\star_1]$. Therefore, $R_1(g \circ \pi^\star_2 \circ h) / R^\star_1 \in [0, 1] \implies C_{\t_1/\t_2} \in [0, 1]$ for any $H, G$.

\textbf{Property \ref{prop:monotonicity}:} Consider $H, H'$ such that $H \subseteq H'$ and $G, G'$ such that $G \subseteq G'$. For any function $f$, the following is true $\forall \pi_2^\star$:
\begin{equation}
    \inf_{h\in H', g\in G'} f(h, g, \pi^\star_2) \leq \inf_{h\in H, g\in G} f(h,g, \pi^\star_2).
\end{equation}  
This implies the following:
% \begin{equation}
%     \inf_{h\in H', g\in G'} \bigg{[}1 - \frac{R_1(g\circ \pi^\star_2 \circ h)}{R^\star_1} \bigg{]} \leq \inf_{h\in H, g\in G} \bigg{[}1 - \frac{R_1(g\circ \pi^\star_2 \circ h)}{R^\star_1} \bigg{]}.
% \end{equation}
% Thus,
\begin{equation}
    \sup_{\pi_2 \in \Pi^\star_2} \inf_{h\in H', g\in G'} \bigg{[}1 - \frac{R_1(g\circ \pi^\star_2 \circ h)}{R^\star_1} \bigg{]} \leq \sup_{\pi^\star_2 \in \Pi^\star_2} \inf_{h\in H, g\in G} \bigg{[}1 - \frac{R_1(g\circ \pi^\star_2 \circ h)}{R^\star_1} \bigg{]}.
\end{equation}

\textbf{Property \ref{prop:relative_complexity 0 iff reduction}:} 
% \edit{Assume $\t_1 \tleq \t_2$. Thus, for all $\pi^\star_2 \in \Pi^\star_2, \ \exists \ g \in G$ and $h \in H$ such that $g\circ \pi^\star_2 \circ h \in \Pi^\star_1$. This implies that for any $\pi^\star_2 \in \Pi^\star_2, \ \exists \ g\in G$ and $h \in H$ such that $R_1(g\circ \pi^\star_2 \circ h) = R^\star_1$. Thus, $C_{\t_1 / \t_2} = 0$}.
Assume $C_{\t_1 / \t_2} = 0$ for some $H$ and $G$  $\iff$ for any $\pi^\star_2 \in \Pi^\star_2 \ \exists \ g\in G$ and $h \in H$ such that $R_1(g\circ \pi^\star_2 \circ h) = R^\star_1$. $R_1(\pi_1) = R^\star_1$ $\iff$ $\pi_1 \in \Pi^\star_1$. Thus, for all $\pi^\star_2 \in \Pi^\star_2 \ \exists \ g \in G$ and $h \in H$ such that $g\circ \pi^\star_2 \circ h \in \Pi^\star_1$ $\iff \t_1 \tleq \t_2$.

\textbf{Property \ref{prop:relative_complexity non0 iff noreduction}:} 
% \edit{The contrapositive of Property \ref{prop:relative_complexity 0 iff reduction} is $C_{\t_1 / \t_2} \neq 0 \implies \neg(\t_1 \tleq \t_2)$. By Property \ref{prop:nonnegative}, the complexity measure is $C_{\t_1 / \t_2} \in [0,1]$, thus $C_{\t_1 / \t_2} \in (0,1] \iff  C_{\t_1 / \t_2} \neq 0$. Thus,  $C_{\t_1 / \t_2} \in (0,1] \implies \neg(\t_1 \tleq \t_2)$.}
The contrapositive of Property \ref{prop:relative_complexity 0 iff reduction} is $\neg(\t_1 \tleq \t_2) \iff C_{\t_1 / \t_2} \neq 0 $. By Property \ref{prop:nonnegative}, the complexity measure is $C_{\t_1 / \t_2} \in [0,1]$, therefore, $C_{\t_1 / \t_2} \in (0,1] \iff  C_{\t_1 / \t_2} \neq 0$. Thus $C_{\t_1 / \t_2} \in (0,1] \iff  \neg(\t_1 \tleq \t_2)$.
\qed} \end{customt}

% \edit{Assume $\neg(\t_1 \tleq \t_2)$. Thus, for some $\pi^\star_2 \in \Pi^\star_2 \ \nexists \ g \in G$ and $h \in H$ such that $g\circ \pi^\star_2 \circ h \in \Pi^\star_1$. This implies that for this $\pi^\star_2 \in \Pi^\star_2$, for any $g\in G$ and $h \in H$, $R_1(g\circ \pi^\star_2 \circ h) < R^\star_1$. Thus, $C_{\t_1 / \t_2} > 0$}. % not true I think now because of for any $g\in G$ and $h \in H$ argument. Can make a sequence of g and h which approaches R^\star_1 and thus C >= 0
\section{Additional Experimental Details}
\label{ap:exp_dets}

\textbf{Approximating Relative Complexity using Q-learning.} We apply Q-learning to Algorithm~\ref{alg:relativecomplexity} by letting the loss functions $L_1$ and $L_2$ correspond to a Q-learning loss: $L_\xi(\pi_\xi) = -\frac{1}{B}\sum_{b=1}^{B}[Q^{\pi_\xi}(s_b,a_b) \log p(a_b)]$, where $p(a_b)$ corresponds to the probability of an action for policy $\pi_\xi$ (which may be a transformation of another policy such as $\pi_\xi = g \circ \pi_{\zeta} \circ h$), $Q^{\pi_\xi}(s_b,a_b)$ are the Q-values, and $B$ is the batch size. We run Algorithm~\ref{alg:relativecomplexity} for 1000 iterations and use a batch size $B$ of 1000 transitions.

\textbf{Approximating Relative Complexity using SAC.} We modify Algorithm~\ref{alg:relativecomplexity} to use SAC for approximating the relative complexity. Let $Q_2^{\pi_2}$ be a critic of $\pi_2$ on task $\t_2$ and $Q_1^{g \circ \pi_2 \circ h}$ be a critic of $g \circ \pi_2 \circ h$ on task $\t_1$. We add an additional step to the algorithm for updating the critics on task $\t_1$ and $\t_2$. The critics are then used in the updates for the policy $\pi_2$ and the encoder/decoder. The resulting method is presented in Algorithm~\ref{alg:relativecomplexitySAC}. We run Algorithm~\ref{alg:relativecomplexitySAC} for 50,000 iterations and use a batch size of 200 transitions. 

\begin{algorithm}[h]
    \caption{Approximating Relative Complexity using SAC}
    \label{alg:relativecomplexitySAC}
\begin{algorithmic}[1]
% \State \textbf{Input:} Threshold for success $R_1^\star, R_2^\star$ on $\t_1, \t_2$ respectively 
\State \textbf{Input:} Learning rates $\lambda_1, \lambda_2$, adversarial tuning parameter $\alpha$
\State \textbf{Input:} Function spaces  $H, G$
\State \textbf{Input:} Q-functions $Q, Q$ loss functions for $\t_1, \t_2$ respectively
\State \textbf{Output:} Approximate relative complexity $\tilde{C}_{\t_1/\t_2} \approx C_{\t_1/\t_2}$
\State \textbf{while} $\neg$(converged $\wedge$ $R_2(\pi_2) = R_2^\star$) \textbf{do}
\State \tab \textbf{Step 0: critic update}
\State \tab Update critic $Q_2^{\pi_2}$ 
\State \tab Update critic $Q_1^{g \circ \pi_2 \circ h}$
\State \tab \textbf{Step 1: $\pi_2$ update} 
\State \tab $\pi_2 \leftarrow \pi_2 + \lambda_1\nabla_{\pi_2}[Q_2^{\pi_2} - \alpha Q_1^{g \circ \pi_2 \circ h}]$
\State \tab \textbf{Step 2: encoder/decoder update} 
\State \tab $[h,g] \leftarrow [h, g] + \lambda_2\nabla_{[h,g]}[Q_1^{g \circ \pi_2 \circ h}]$
\State \textbf{end while}
\State $\tilde{C}_{\t_1/\t_2} \leftarrow \bigg{[}1 - \frac{R_1(g\circ \pi^\star_2 \circ h)}{R_1^\star} \bigg{]}$ 
\end{algorithmic}
\end{algorithm}

\end{document}